\documentclass[sigconf]{acmart}
\usepackage{nicefrac}
\usepackage{siunitx}
\usepackage{array,framed}
\usepackage{booktabs}
\usepackage{pifont}
\newcommand{\cmark}{\ding{51}}%
\newcommand{\xmark}{\ding{55}}%
\usepackage{
  color,
  float,
  epsfig,
  wrapfig,
  graphics,
  graphicx,
  subcaption
}

\usepackage{textcomp,amssymb}
\usepackage{setspace}
\usepackage{latexsym,fancyhdr,url}
\usepackage{enumerate}
\usepackage{algorithm2e}
\usepackage{algpseudocode}
\usepackage{graphics}
\usepackage{xparse} 
\usepackage{xspace}
\usepackage{multirow}
\usepackage{csvsimple}
\usepackage{balance}

\usepackage{
  tikz,
  pgfplots,
  pgfplotstable
}
\usepackage{hyperref}

\usetikzlibrary{
  shapes.geometric,
  arrows,
  external,
  pgfplots.groupplots,
  matrix
}

\pgfplotsset{compat=1.9}

\usepackage{mathtools,}

\DeclareMathAlphabet{\mathcal}{OMS}{cmsy}{m}{n}

\DeclareGraphicsExtensions{%
    .png,.PNG,%
    .pdf,.PDF,%
    .jpg,.mps,.jpeg,.jbig2,.jb2,.JPG,.JPEG,.JBIG2,.JB2}

\usepackage{xparse}
\newcommand{\bnm}{\begin{newmath}}
\newcommand{\enm}{\end{newmath}}

\newcommand{\bea}{\begin{eqnarray*}}%
\newcommand{\eea}{\end{eqnarray*}}%

\newcommand{\bne}{\begin{newequation}}
\newcommand{\ene}{\end{newequation}}

\newcommand{\bal}{\begin{newalign}}
\newcommand{\eal}{\end{newalign}}

\newenvironment{newalign}{\begin{align}%
\setlength{\abovedisplayskip}{4pt}%
\setlength{\belowdisplayskip}{4pt}%
\setlength{\abovedisplayshortskip}{6pt}%
\setlength{\belowdisplayshortskip}{6pt} }{\end{align}}

\newenvironment{newmath}{\begin{displaymath}%
\setlength{\abovedisplayskip}{4pt}%
\setlength{\belowdisplayskip}{4pt}%
\setlength{\abovedisplayshortskip}{6pt}%
\setlength{\belowdisplayshortskip}{6pt} }{\end{displaymath}}

\newenvironment{newequation}{\begin{equation}%
\setlength{\abovedisplayskip}{4pt}%
\setlength{\belowdisplayskip}{4pt}%
\setlength{\abovedisplayshortskip}{6pt}%
\setlength{\belowdisplayshortskip}{6pt} }{\end{equation}}

\newcounter{ctr}

\newcounter{mytable}
\def\mytable{\begin{centering}\refstepcounter{mytable}}
\def\endmytable{\end{centering}}

\newcounter{myfig}
\def\myfig{\begin{centering}\refstepcounter{myfig}}
\def\endmyfig{\end{centering}}

\newlength{\saveparindent}
\newlength{\saveparskip}
\newcommand{\E}{{\rm I\kern-.3em E}}

\renewcommand{\eqref}[1]{\mbox{Equation~(\ref{#1})}}

\def \part {part}

\renewcommand{\paragraph}[1]{\vspace*{6pt}\noindent\textbf{#1}\;}

\def \blackslug{\hbox{\hskip 1pt \vrule width 4pt height 8pt
    depth 1.5pt \hskip 1pt}}
\def \qed{\quad\blackslug\lower 8.5pt\null\par}

\newcounter{mynote}[section]

\newcommand\ignore[1]{}

\newcounter{rcnote}[section]

\newcounter{mrnote}[section]

\newcounter{fknote}[section]

\newcounter{anote}[section]

\DeclareMathSymbol{\mlq}{\mathord}{operators}{``}
\DeclareMathSymbol{\mrq}{\mathord}{operators}{`'}

\newcommand{\rhf}[2]{R_{f, \gamma}}

\DeclareDocumentCommand{\edist}{o o}{
  \ensuremath{
    \IfNoValueTF{#1}{{d}}{{\sf d}(#1,#2)}
  }
}

\newcommand{\olrk}[1]{\ifx\nursymbol#1\else\!\!\mskip4.5mu plus 0.5mu\left(\mskip0.5mu plus0.5mu #1\mskip1.5mu plus0.5mu \right)\fi}

\NewDocumentCommand{\indseq}{ O{1} O{r} }{{#1}\ldots {#2}}

\setcopyright{none}
\begin{document}
\fancyhead{}

\title{EEG-based Cognitive Load Classification using Feature Masked Autoencoding and Emotion Transfer Learning}

\author{Dustin Pulver, Prithila Angkan, Paul Hungler, Ali Etemad}
\affiliation{
  \institution{Queen's University, Canada}
  \country{}
}

\renewcommand{\shortauthors}{Trovato and Tobin, et al.}

\begin{abstract}
Cognitive load, the amount of mental effort required for task completion, plays an important role in performance and decision-making outcomes, making its classification and analysis essential in various sensitive domains. In this paper, we present a new solution for the classification of cognitive load using electroencephalogram (EEG). Our model uses a transformer architecture employing transfer learning between emotions and cognitive load. We pre-train our model using self-supervised masked autoencoding on emotion-related EEG datasets and use transfer learning with both frozen weights and fine-tuning to perform downstream cognitive load classification. To evaluate our method, we carry out a series of experiments utilizing two publicly available EEG-based emotion datasets, namely SEED and SEED-IV, for pre-training, while we use the CL-Drive dataset for downstream cognitive load classification. The results of our experiments show that our proposed approach achieves strong results and outperforms conventional single-stage fully supervised learning. Moreover, we perform detailed ablation and sensitivity studies to evaluate the impact of different aspects of our proposed solution. This research contributes to the growing body of literature in affective computing with a focus on cognitive load, and opens up new avenues for future research in the field of cross-domain transfer learning using self-supervised pre-training.
\end{abstract}



\begin{teaserfigure}
  \includegraphics[width=\textwidth]{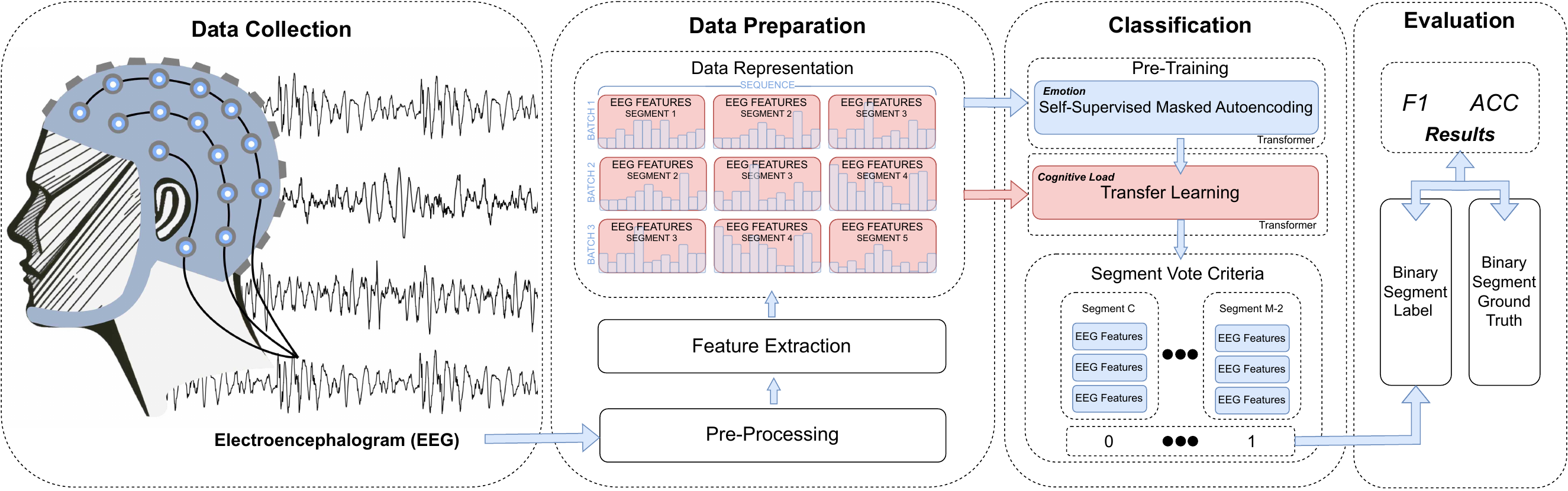}
  \caption{A complete overview of the pipeline used in our paper is presented.}
  \label{fig:teaser}
  \vspace{10mm}
\end{teaserfigure}




\maketitle

\section{Introduction}
Cognitive load refers to the amount of mental effort required to process information and complete a task. It is a crucial concept in educational psychology \cite{sweller2016cognitive}, human-computer interaction \cite{hollender2010integrating}, healthcare \cite{fraser2015cognitive}, as well as other domains concerned with learning and information processing. The concept of cognitive load has been used to explain how people learn, perform tasks, and interact with technology. The level of cognitive load experienced by an individual performing a task can significantly influence their success or failure in performance \cite{das2014cognitive}. This makes the ability to automatically classify cognitive load extremely important in a variety of sensitive domains. For instance, in healthcare, the cognitive load of medical professionals can affect their decision-making and patient outcomes \cite{ehrmann2022evaluating}. In aviation, pilots' cognitive load can affect their ability to respond to unexpected events and make critical decisions \cite{huttunen2011effect}. In the context of driving, prior research has investigated the relationship between cognitive load and driving performance, suggesting that an excessive amount of cognitive load may elevate the likelihood of road accidents \cite{engstrom2017effects}. 

Many approaches have been presented to determine one's cognitive load including the use of speech \cite{hecker2022quantifying}, vision \cite{biondi2023distracted}, and bio-signals such as EEG \cite{antonenko2010using}. This paper further explores the use of EEG to detect cognitive load. EEG is a non-invasive technique used to record the electrical activity of the brain with high temporal resolution. These signals are generated by the synchronous firing of groups of neurons and are recorded by placing electrodes on the scalp which detect the voltage fluctuations resulting from neural activity \cite{gloor1969hans}. The resultant signal encompasses a succession of waves exhibiting diverse frequencies and amplitudes, which are indicative of distinct cognitive and physiological states. Numerous deep learning techniques have been employed for EEG analysis, resulting in significant advancements in the field \cite{zhang2020spatio, zhang2019classification, zhang2023partial}.

The investigation of cognitive load remains a relatively unexplored area of research but has recently gained momentum, particularly with the development of comprehensive cognitive load datasets in different contexts such as detecting passenger cognitive states \cite{angkan2023multimodal}. The detection and quantification of cognitive load among vehicle passengers could enhance their safety, comfort, and overall experience. In order to effectively learn EEG data, both spatial and temporal relationships need to be captured. Transformers \cite{vaswani2017attention} have recently emerged as a powerful tool, capable of learning both spatial and temporal information, and have shown impressive performances in a variety of different domains, such as natural language processing \cite{devlin2018bert}, computer vision \cite{dosovitskiy2020image}, and healthcare \cite{gupta2017towards}. Transformers, however, require large amounts of data for effective training, making them difficult to use for applications such as EEG-based cognitive load analysis, for which few datasets exist.

To tackle the problem mentioned above we propose a transformer architecture for classifying cognitive load based on EEG signal. In order to overcome the data scarcity issue, we explore transfer learning between EEG-based emotion datasets (namely SEED \cite{duan2013differential, zheng2015investigating} and SEED-IV \cite{zheng2018emotionmeter}) and a cognitive load dataset (namely CL-Drive \cite{angkan2023multimodal}). We pre-train the transformer model using self-supervised masked autoencoding on the emotion datasets, following freezing of the transformer blocks, we transfer the model for downstream cognitive load classification on CL-Drive. In this stage, we keep the transformer blocks frozen, and only train the classification head. Our results demonstrate strong performance in comparison to fully supervised training directly on the downstream cognitive load dataset, demonstrating the important potential for emotion datasets to be used for cognitive load pre-training.

The contributions of this paper are summarized as follows:
\begin{itemize}

\item For the first time, we perform EEG-based cognitive load classification using masked autoencoding of features and a transformer architecture, representing a previously unexplored approach in this domain.

\item Our method uses transfer learning between emotions and cognitive load using EEG signals. We pre-train our model using self-supervised masked autoencoding on emotion-related EEG datasets and use transfer learning with frozen weights to perform downstream cognitive load classification.

\item Our method achieves strong results on cognitive load classification. Moreover, our results suggest that transfer learning between emotions and cognitive load is indeed a viable path for cognitive load analysis with deep learning given that our approach outperforms the conventional single-stage fully supervised learning.

\end{itemize}

The rest of this paper is organized as follows. In Section~\ref{Related Work}, we present a summary of EEG learning with transformer architectures and discuss the application of self-supervised pre-training to various bio-signals, including EEG. The proposed methodology is outlined in Section~\ref{Method}, which is illustrated in Figure~\ref{fig:teaser}. We then describe the experimental setup, including the datasets, evaluation protocol, and implementation details, in Section~\ref{Experimentation}. In Section~\ref{results}, we present the experimental results and report the conducted ablation studies and sensitivity analysis. Finally, we conclude the paper and suggest future works in Section~\ref{Conclusion&FutureWork}.

\section{Related Work} \label{Related Work}
In this section we first explore the use of transformer architectures for learning EEG signals, followed by an investigation into the promising technique of self-supervised pre-training and their use with bio-signals. Furthermore, we provide a more in-depth examination of cross-domain transfer learning.

\subsection{EEG Learning with Transformers}
Since their introduction in \cite{vaswani2017attention}, transformers have outperformed many state-of-the-art models on a variety of different sequence modeling tasks, including machine translation \cite{vaswani2017attention,wang2019learning}, language modeling \cite{radford2018improving,devlin2018bert,brown2020language,vaswani2017attention}, and image captioning \cite{anderson2018bottom,xu2015show}. The attention mechanism in transformers allows the model to attend to different parts of the input sequence, capturing long-range dependencies and improving performance.  
Recently, transformers have been adopted for use in the domain of bio-signals \cite{behinaein2021transformer}, including for analysis of EEG data. In \cite{tao2020eeg}, an approach was proposed for EEG-based emotion recognition using channel-wise attention and self-attention mechanisms. Combining transformer architectures with convolutional neural networks (CNN) has shown great promise for EEG signal recognition, as demonstrated by recent research \cite{sun2021eeg}. Despite their promising performance in other domains, the adoption of transformers has been less prominent for bio-signals like EEG. This can be attributed to the substantial amount of training data required to effectively train transformers, which may not always be feasible for bio-signals such as EEG. More specifically, to the best of our knowledge, the current literature has yet to explore the use of transformer architectures for EEG-based cognitive load classification.

\subsection{Self-Supervised Pre-Training}
Recent advancements in self-supervised learning have significantly enhanced the effectiveness of models across various domains such as language \cite{devlin2018bert,brown2020language}, vision \cite{chen2020simple,chen2021exploring}, and bio-signals \cite{sarkar2020self,sarkar2021detection}. The prevalent approach in this field involves utilizing self-supervision for pre-training, subsequently fine-tuning the model for downstream classification tasks, or maintaining it in a frozen state. Recently, masked autoencoding using transformers \cite{he2022masked} has emerged as an effective and scalable representation learning framework. This approach incorporates masked autoencoders in a self-supervised fashion, where random segments of the input image are concealed, and an encoder-decoder configuration is used to reconstruct the missing pixels. Self-supervised techniques, such as masked autoencoding, have been extensively employed in conjunction with features (rather than raw data) across various domains \cite{wei2022masked,yan20233d,jing2020self}.

Recently, the efficacy of pre-training techniques has been explored in the context of EEG signal analysis. For instance, in \cite{chien2022maeeg} self-supervised masked autoencoding has been utilized for EEG sleep stage classification. For pre-training, their approach involves extracting features from raw EEG signals with the use of a CNN, followed by a transformer encoder tasked with reconstructing partially masked features. Following the encoder, two additional layers map the model output to the dimension of the raw EEG signal to compute a reconstructive loss. For fine-tuning, an additional linear classification layer is added to perform a variety of downstream tasks. A similar approach was proposed in \cite{kostas2021bendr}, where a contrastive loss was calculated by directly comparing the encoder output with CNN output features. The results of \cite{chien2022maeeg} and \cite{kostas2021bendr} demonstrate the effectiveness of self-supervised learning approaches for EEG learning, suggesting their applicability to other classification tasks, including cognitive load classification.

Lastly, the notion of cross-domain transfer learning has been recently explored for EEG representation learning. In \cite{yang2023cross}, cross-domain transfer learning is applied between patient-specific seizure prediction and sleep staging. First, a simple CNN was trained to conduct the cross-domain prediction tasks in both directions, seizure to sleep and vice versa. After pre-training the model for one specific task, 6 layers are then frozen to enable cross-domain transfer for the other task. Recently, in \cite{li2022spp}, the use of self-supervised cross-domain transfer learning was investigated for EEG learning. Their approach implements a CNN pre-trained with a contrastive self-supervised learning task. For pre-training a clinical EEG dataset labeled as normal or abnormal was used, although the labels were excluded for the self-supervised task. For the downstream task, multiple linear layers were added to the CNN as a prediction head to classify EEG binary motor imagery tasks. 

\section{Method}\label{Method}

The goal of this paper is to effectively classify cognitive load from EEG signals using transformer architectures. However, our initial experiments suggest that conventional fully supervised single-stage training is not sufficiently effective for this task. Consequently, we investigate self-supervised pre-training using cross-domain datasets due to the lack of EEG cognitive load data sources. Our proposed method exploits the use of cross-domain transfer learning to effectively classify cognitive load from EEG signals. Next, we describe the pipeline of our method, beginning with data tokenization, followed by model pre-training, and downstream cognitive load classification.

\begin{figure}[t]
\centering
\includegraphics[width=1\columnwidth]{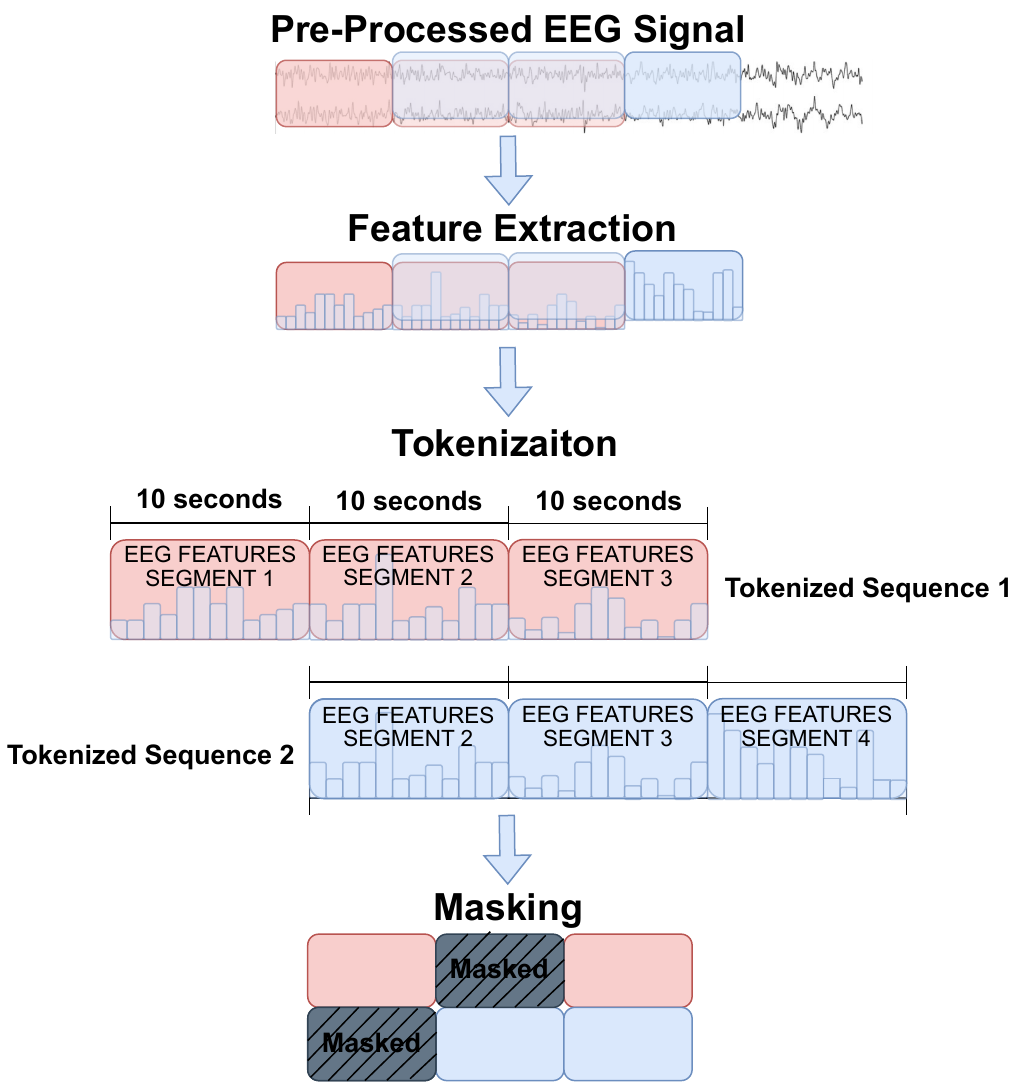}
\caption{The process of converting raw EEG signals into tokenized sequences of EEG features is presented.}
\label{figToken}
\end{figure}

\subsection{Tokenization}

To enhance the quality of the EEG recordings, we first apply pre-processing to all the datasets used for both pre-training and downstream cognitive load classification (see Section~\ref{Datasets}). Specifically, we apply a 2nd order Butterworth bandpass filter with a passband frequency of 1-75 Hz to eliminate unwanted noise and artifacts. Additionally, a notch filter with a quality factor of 30 was utilized to remove powerline noise at a frequency of 60 Hz.

We then conduct feature extraction from all the datasets used in this study. Following prior works \cite{zhang2020spatio,zhang2022parse,angkan2023multimodal,wu2019identifying}, two key features namely power spectral density (PSD) and Differential entropy (DE), are extracted. Both features were extracted over 5 frequency bands, from 1 to 75 Hz, namely, Delta (1-4 Hz), Theta (4-8 Hz), Alpha (8-12 Hz), Beta (12-31 Hz) and Gamma (31-75 Hz), over a 10-second window. PSD and DE are considered to encompass a large portion of the necessary information from raw EEG signals as they capture important characteristics such as the distribution of power across different frequency bands and the complexity of the signal, and are thus widely used as the main features in the area. 

Effectively, given pre-processed EEG signals $X \in \mathbb{R}^{b \times c}$, where $X$ are the EEG signal values, $b$ is the number of bands, and $c$ is the number of channels. Successive to feature extraction, we have $X_{PSD} \in \mathbb{R}^{b \times c}$ and $X_{DE} \in \mathbb{R}^{b \times c}$. Given our focus on 5 frequency bands and the use of 4 channels (as will be discussed in Section~\ref{sensorPlacement}), $b = 5$ and $c = 4$ respectively. PSD is a measure of signal power across different frequency components. To calculate this feature, we utilize Welch's method for each frequency band \cite{solomon1991psd}. The Welch method segments the signal into smaller sections, applies a window function to each segment, computes the discrete Fourier Transform of each segment, and then averages their squared magnitudes. This approach effectively reduces noise and provides a more accurate representation of the power spectrum across different frequency bands. Differential entropy (DE) is a measure of the complexity or irregularity of an EEG signal, and it is derived from the concept of differential entropy in information theory \cite{duan2013differential}. To calculate DE, we assume that the EEG signal has a Gaussian distribution.

Next, we concatenate and z-score normalize $X_{PSD}$ and $X_{DE}$ to obtain $X_{feat} \in \mathbb{R}^{2 \times b \times c}$. We then tokenize $X_{feat}$ into smaller, more manageable sequences, which can be processed and analyzed more efficiently by the model. Figure~\ref{figToken} depicts the tokenization process. We choose 10-second segments with no overlap as our downstream cognitive load dataset (see Section~\ref{Datasets}) provide output labels for every 10-second window. Let $X^{i}_{feat}$ be the set of features for the $i$th 10-second window, where $i = 1, 2, ..., n$, and $n$ is the number of 10-second segments (with no overlap) available in each participant's trial. Accordingly, we define a tokenized sequence as $S^j = [X^{i}_{feat}, X^{i+1}_{feat}, X^{i+2}_{feat}]$, where $S^{j}$ is the $j$th sequence of features, and $j= 1, 2, ..., n-2$. Now that the data is arranged in sequences, it can be used to train the transformer architecture found in our proposed approach in the following sections.

\subsection{Masked Autoencoding and Transfer Learning}
\begin{figure}[t]
\centering
\includegraphics[width=0.75\columnwidth]{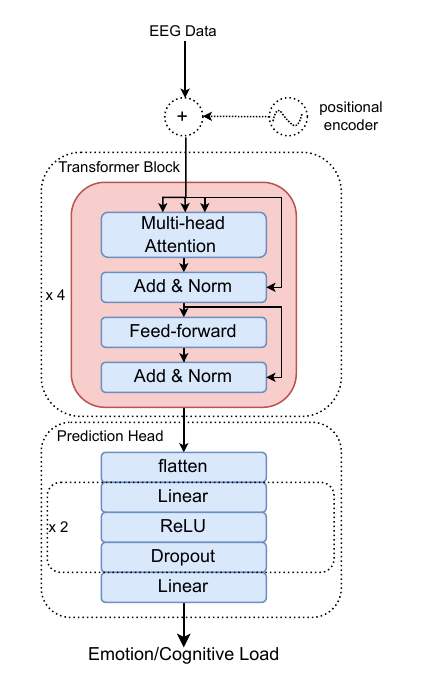}
\caption{A diagram of the transformer architecture used in our solution is presented.}
\label{figARC}
\end{figure}

In the pre-training stage, we conduct self-supervised masked autoencoding on the tokenized sequences of EEG feature segments, $S^j$. Our model consists of a masked self-supervised autoencoder,  which learns to reconstruct partially masked sequences of features during the training process. To create these masked sequences, we randomly mask one of the three segments in each tokenized sequence $S_j$, which we denote by $g_j$. 
The model consists of a transformer encoder comprising 4 transformer blocks, followed by a prediction head. The first component of the prediction head is a flattening layer that flattens the output of the transformer encoder. This is followed by 2 prediction blocks consisting of FC layers followed by ReLU activation with dropout. The size of the FC layers are 256 and 128 respectively. This is followed by a linear layer with size equal to the dimension of $g_j$. The architecture of our model is depicted in Figure~\ref{figARC}. The model is tasked with predicting the features of the masked segment using information from the remaining unmasked segments in the sequence. During the pre-training stage, the model is tasked with performing regression on the features of the masked segment. We feed the masked sequences to the model, and train it using the L1 loss function measuring the mean absolute error (MAE) between the ground truth masked segments, $g_j$, and their predicted values $\hat{g_j}$, as follows:
\begin{equation}\label{L1}
L_{1}(g_j, \hat{g_j}) = \frac{1}{M} \sum_{i=1}^{M} | g_j - \hat{g_j} |,
\end{equation}
where $M$ refers to the number of masked segments. 

Following pre-training, we aim to use the model for downstream EEG cognitive load classification. To this end, we discard the prediction head and transfer the transformer blocks. In order to enable the model to classify cognitive load, we incorporate a new head consisting of a flattening layer that flattens the output of the transformer encoder, followed by 2 prediction blocks comprising an FC layer followed by ReLU activation with dropout. The size of the FC layers are 32 and 16, respectively. This is followed by a linear layer with the size of 3, and an additional Sigmoid activation function used to output the predicted binary cognitive load values for each 10-second segment $X^{i}_{feat}$ of $S_j$. To investigate the effectiveness of the pre-trained model used in the downstream cognitive load classification task, we explore two distinct approaches. The first approach involves using the transformer blocks in a frozen fashion and only training the prediction head, while the latter allows them to be trained along with the new prediction head. In both cases, we train the model (either only the prediction head or the entire model) using the Binary Cross Entropy loss function measuring the difference between the predicted binary cognitive load labels $\hat{y}$ and the ground truth labels $y$, as follows:
\begin{equation}\label{BCE}
\mathrm{BCE}(y,\hat{y}) = -\frac{1}{C}\sum_{i=1}^{C}[y_i\log{\hat{y}_i} + (1-y_i)\log{(1-\hat{y}_i)}],
\end{equation}
where $C$ refers to the number of segments. 

\subsection{Segment Vote Aggregation}
Once the model is ready for the downstream classification task, it will generate a sequence of binary cognitive load predictions for each segment $X^{i}_{feat}$ in the tokenized input sequence $S^{j}$. Consequently, due to the overlapping sequences (see Figure~\ref{figToken}), each segment will be present in three consecutive sequences, and therefore result in 3 separate predictions. Thus, an aggregation process is necessary to determine a final label for each segment. To generate a predicted value for each segment, we use a voting mechanism among the three predicted labels for that segment. It should be noted that the first two and the last two segments in each batch will not have 3 predictions on which to vote, and therefore those segments are excluded from evaluation.

\section{Experiments}\label{Experimentation}
Here, we provide the details on the experiments carried out in this study. First, we describe the emotion datasets used during pre-training along with the dataset used for downstream cognitive load classification. These datasets have been collected with differing numbers of EEG channels. We address the discrepancy between the datasets and explain how we mitigated it. Next, we outline the evaluation protocol employed to assess the performance of our approach. Lastly, we present the implementation details required to replicate our experiments.

\subsection{Datasets} \label{Datasets}
\subsubsection{Pre-Training (Emotion) Datasets}
Two datasets, SEED \cite{duan2013differential, zheng2015investigating} and SEED-IV \cite{zheng2018emotionmeter}, are used for pre-training, both individually as well as combined. These datasets are both emotion-related and commonly used for affective computing. As the pre-training task is self-supervised, the labels of these datasets will not be used.

\noindent The \textbf{SEED \cite{duan2013differential, zheng2015investigating}} dataset consists of 15 film clips used as stimuli during data collection. Each clip is categorized into 3 class labels (positive, neutral, and negative emotions). The data collection involved presenting each film clip for 4 minutes, with each of the 15 participants repeating this process twice. Of the 15 participants 7 were male and 8 were female, with an average age of 23.27. The EEG signals of each participant have been recorded using 62 channels at a 1000 Hz sampling rate, which was later downsampled to 200 Hz.

\noindent \textbf{SEED-IV \cite{zheng2018emotionmeter}} consists of 72 short film clips used as stimuli during data collection. Each clip is categorized into 4 class labels (happy, sad, neutral, and fear emotions). The data collection involved presenting 24 short film clips at each session. Of the 15 subjects who participated in this study, 7 were male and 8 were female, all between the ages of 20-24. With each participant completing 3 sessions each on separate days, with their EEG signals recorded using 62 channels at a 1000 Hz sampling rate. The signals were later downsampled to 200 Hz.

\begin{figure}[t]
\centering
\includegraphics[width=0.65\columnwidth]{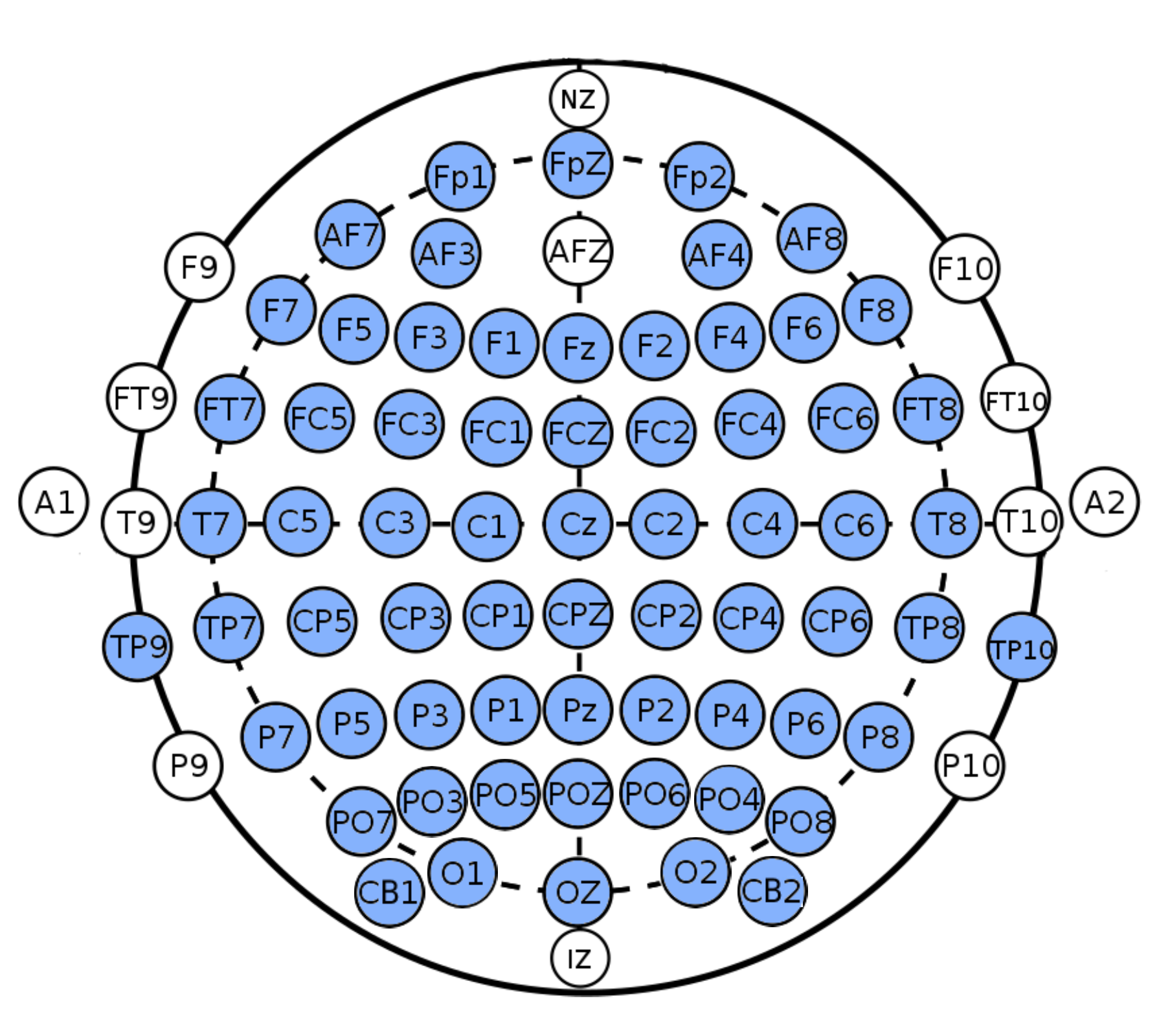}
\caption{The 10-20 system for electrode placements used by the SEED and SEED-IV datasets.}
\label{figSEED}
\end{figure}

\begin{figure}[t]
\centering
\includegraphics[width=0.65\columnwidth]{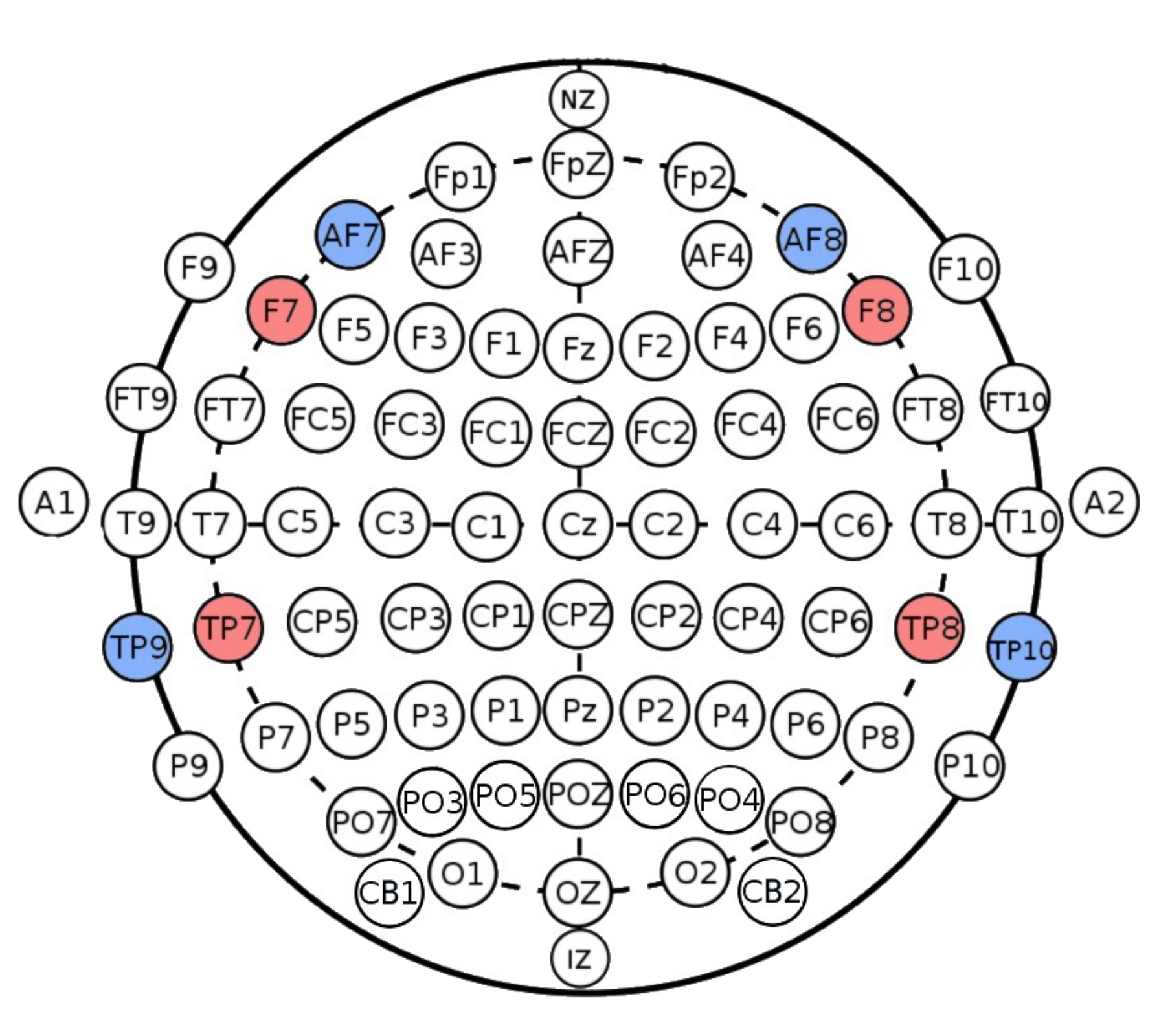}
\caption{Comparison of the electrode placements in the CL-DRIVE dataset (blue) to the selected electrodes in the SEED and SEED-IV datasets (red), using the 10-20 system.}
\label{figCOMPARE}
\end{figure}

\subsubsection{Downstream (Cognitive Load) Dataset}
We use \textbf{CL-Drive \cite{angkan2023multimodal}} as the downstream dataset for cognitive load classification. CL-Drive is a 
driver cognitive load assessment dataset, containing a number of different physiological signals including EEG. The data was collected in a physical vehicle driving simulator, where 23 participants completed 3-minute tasks ranging over 9 complexity levels. Of the 23 participants, 6 were male and 17 were female. During the tasks, the participants reported their subjective cognitive load every 10 seconds on a scale from 1-9, thus providing cognitive load labels for each 10-second segment of EEG data. The categorical labels were transformed into binary values, with values in the range of 1 to 5 representing low cognitive load (0), and values in the range of 6 to 9 representing high cognitive load (1). The EEG signals were recorded using a 4-channel wearable headband at a 256 Hz sampling rate.

\subsubsection{Sensor placement}\label{sensorPlacement}
All three datasets have been recorded using EEG sensor setups that follow the 10-20 system for electrode placement \cite{morley201610}. This system is widely used for consistent EEG electrode placement across individuals and studies, aiding in data consistency. A diagram of the electrode placements for both the SEED and SEED-IV datasets can be seen in Figure~\ref{figSEED}. The CL-Drive dataset and the SEED datasets differ in their electrode placements, with the former using only 4 electrodes, at locations, TP9, AF7, AF8, and TP10, while the latter consists of 62 electrodes. To address this difference, we selected four electrodes from the SEED datasets that closely matched the placements in the CL-Drive dataset. Figure~\ref{figCOMPARE} shows the most similar electrode pairs, with the first element of each pair being from the SEED datasets (depicted in Red) and the second from the CL-Drive dataset (depicted in Blue), namely (TP7, TP9), (F7, AF7), (F8, AF8), and (TP8, TP10). 

\subsection{Evaluation Protocol}
To evaluate our proposed pipeline, we follow the evaluation protocol used for cognitive load classification in the paper that presents the CL-Drive dataset \cite{angkan2023multimodal}. A 10-fold cross-validation is performed on the downstream cognitive load classification. Given the imbalance between low and high cognitive load data (38\% high cognitive load and 62\% low cognitive load), we report both macro F1 scores as well as accuracy over the segment vote criteria for each fold.

\subsection{Implementation Details}\label{Implementation Details}
We provide a description of the specific parameters that are used to implement our proposed method. For both pre-training and downstream classification, a batch size of 64 and the Adam optimizer were used. The learning rate for the Adam optimizer was held constant at 0.0001 during pre-training, but a dynamic learning rate was applied during the downstream task, where fine-tuning was applied. We also experimented with learning rate schedulers, the results for which are presented in the next sections of the paper. During both the pre-training and downstream stages, we trained the model over 1000 epochs.

\section{Results}\label{results}     
In this section, we conduct a series of experiments to evaluate the proposed approach for EEG cognitive load classification. First, we explore the impact of fine-tuning the entire model on the downstream task vs. keeping the transformer blocks frozen. We then evaluate the impact of using PSD, DE, or both, and also further experiment with fully-supervised training as a baseline. Then we perform ablation experiments followed by sensitivity studies on the important components and parameters of our solution. The numerical values presented in brackets beside each score denote the standard deviation of the respective values. In every table, the best score is highlighted in bold.

\subsection{Performance}\label{FrozenVFine}

We investigate the effect of fine-tuning the pre-trained encoder on the performance of the downstream EEG cognitive load classification task. The results of these experiments are summarized in Table~\ref{table:Frozen}, where we use PSD features from the SEED, SEED-IV, and a combination of the two datasets to pre-train the encoder. Our results demonstrate that freezing the pre-trained transformer architecture and only training the prediction head is more effective than fine-tuning the entire model. We believe one explanation for this observation is that when fine-tuned, the model is not able to retain the knowledge learned from the pre-training stage. This itself could be due to overfitting, which could occur when the downstream target dataset is relatively small in size. Research has demonstrated that the decision to fine-tune a downstream task hinges on the size of the target dataset, as evidenced in \cite{yosinski2014transferable}. On the other hand, when the transformer blocks are kept frozen, given the relative similarity of emotions and cognitive load \cite{hunziker2021situated}\cite{plass2019four}, the model is able to retain and use the learned knowledge in the pre-training stage to extract effective representations for the downstream classification task. 

\begin{figure}[t]
\centering
\includegraphics[width=0.9\columnwidth]{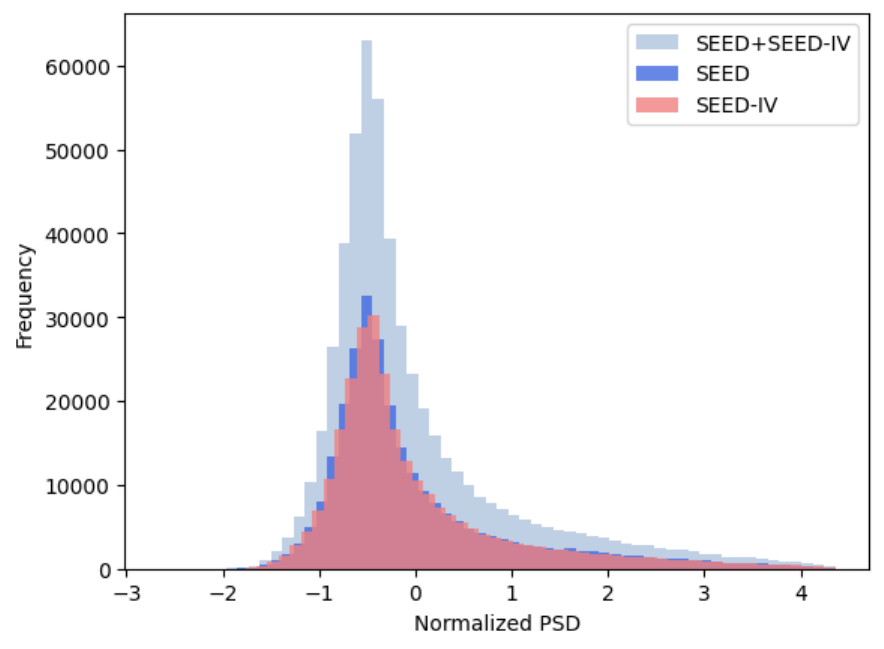}
\caption{The distributions of the datasets used in pre-training.}
  \vspace{8mm}
\label{figDIST}
\end{figure}

In Table~\ref{table:Frozen}, we investigate the impact of pre-training datasets on downstream task performance and observe that while using both SEED and SEED-IV is slightly better in performance than using either of them alone, the difference is not significant as SEED, SEED-IV, and SEED + SEED-IV yield relatively similar downstream results despite the expectation that a larger pre-training dataset would enhance performance. To further explore the reason behind this, we analyze the distributions of these datasets and present the outcome in Figure~\ref{figDIST}. We observe in this figure that the distributions of SEED and SEED-IV are almost identical, which explains the reason for the lack of meaningful changes in downstream performance. 

  \begin{table}[t]
        \caption{Cognitive load classification results using frozen and fine-tuning strategies, with pre-training on SEED, SEED-IV, and SEED+SEED-IV.}
        \label{table:Frozen}
        \small
    \setlength
    \tabcolsep{2pt}
        \begin{center}{
            \begin{tabular}{l|ll|ll}
                 \hline
            \multicolumn{1}{c|}{} & \multicolumn{2}{|c}{Frozen}& \multicolumn{2}{|c}{Fine-Tuned}\\
                 \hline
                 Datasets & Accuracy & F1 score &  Accuracy & F1 score\\
                 \hline\hline
                 SEED & 73.57 (0.016) & 68.99 (0.026) & 67.94 (0.021) & 60.35 (0.043) \\
                 SEED-IV & 73.81 (0.029) & 69.59 (0.031) & 68.26 (0.043) & 57.60 (0.067)  \\
                 SEED + SEED-IV & \textbf{74.07 (0.025)} & \textbf{70.28 (0.025)} & 63.06 (0.028) & 52.75 (0.070)  \\
                 \hline
                \end{tabular} 
                }
        \end{center}
\end{table}

To further evaluate the impact of pre-training, we compare the performance of our solution pre-trained on SEED and SEED-IV to a fully supervised setup where no pre-training is performed. Moreover, we experiment with using PSD, DE, and PSD+DE as the input representations. The results of these experiments are presented in Table~\ref{table:data-pre-train}, which demonstrate a significant improvement in performance when utilizing pre-training in comparison to fully supervised learning without the use of pre-training. An average increase of 8.33\% in accuracy and 11.33\% in F1 is observed across all experiments. We believe this performance increase can be attributed to the additional information gained by the model during the self-supervised pre-training with masked autoencoding. Moreover, we observe that the PSD features alone provide a better input representation for cognitive load classification in comparison to DE or PSD+DE.

 \begin{table}[t]
        \caption{The impact of pre-training vs. fully supervised learning, as well as the impact of features used.}
        \label{table:data-pre-train}
        \small
    \setlength
    \tabcolsep{3pt}
        \begin{center}{
                \begin{tabular}{l|ll|ll}
                 \hline
                 \multicolumn{1}{c|}{} & \multicolumn{2}{|c}{Pre-Training}& \multicolumn{2}{|c}{Fully Supervised}\\
                 \hline
                 Features & Accuracy & F1 score &  Accuracy & F1 score\\
                 \hline\hline
                 PSD & \textbf{74.07 (0.025)} & \textbf{70.28 (0.025)} & 64.92 (0.054) & 56.87 (0.029)\\
                 DE & 71.67 (0.031) & 68.57 (0.038) & 65.55 (0.033) & 52.06 (0.076)\\
                 PSD + DE & 70.69 (0.043) & 67.68 (0.035) & 67.27 (0.039) & 56.72 (0.058)\\
                 \hline
                \end{tabular} 
                }
        \end{center}
\end{table}

\begin{table}[t]
        \caption{Ablation experiments on the transformer blocks.}
        \label{table:blocks}
        \small
        \begin{center}{
        \begin{tabular}{lll}
                 \hline
                 Blocks & Accuracy & F1 score  \\
                 \hline\hline
                 5  & 72.45 (0.023) & 69.76 (0.019) \\
                 4  & \textbf{74.07 (0.025)} & \textbf{70.28 (0.025)} \\
                 3  & 72.18 (0.030) & 68.01 (69.43) \\
                 \hline
                \end{tabular} 
                }
        \end{center}
\end{table}

\begin{table}[t]
        \caption{Ablation experiments on the prediction head of the cognitive load classifier.} 
        \label{table:predHeadAbl} 
        \small
        \begin{center}{
                \begin{tabular}{lll}
                 \hline
                 Arc. Name & Accuracy & F1 score  \\
                 \hline\hline
                 A1  & 74.07 (0.025) & 70.28 (0.025) \\
                 A2 & \textbf{74.55 (0.027)} & \textbf{71.17 (0.031)} \\
                 A3  & 74.32 (0.021) & 69.95 (0.035) \\
                 \hline
                \end{tabular} 
                }
        \end{center}
\end{table}

\begin{table}[t]
        \caption{Ablation experiment on the impact of positional encoding.} 
        \label{table:positional}
        \small
        \begin{center}{
        \begin{tabular}{lll}
        \hline
        Positional Enc. & Accuracy & F1 score  \\
        \hline\hline
        \cmark  & 70.87 (0.024) & 66.93 (0.023) \\
        \xmark & \textbf{74.07 (0.025)} & \textbf{70.28 (0.025)} \\
        \hline
        \end{tabular} 
        }
        \end{center}
\end{table}

\subsection{Ablation Studies}\label{Ablation} 

Here, we ablate the key components of our method to investigate their importance and contribution towards the final performance. To this end, we explore the impact of downstream prediction head architecture, number of transformer blocks, and the use of positional encoding. To standardize these studies we pre-trained on the combination of SEED and SEED-IV datasets and only used PSD features.

First, we ablate the number of transformer blocks and explore 5, 4, and 3 blocks respectively. The results of this study can be seen in Table~\ref{table:blocks}, which demonstrate that the use of 4 transformer blocks achieves the best performance. 

Next, we ablate the layers of the downstream prediction head. 3 different prediction head architectures were tested in this ablation study to determine the most effective parameter choices. The first architecture includes 3 FC layers, with sizes 32, 13, and 8 respectively, which we refer to as $A1$. The second architecture includes 2 FC layers, with sizes 32 and 13, respectively, named $A2$. The last architecture includes 1 FC layer with a size of 32, which we refer to as $A3$. All three architectures are followed by a linear layer with a size of 3. As per the previous ablation study, we use 4 transformer blocks in all three variants. The results of this study are presented in Table~\ref{table:predHeadAbl}, which demonstrates prediction head architecture $A2$ as the ideal choice. 

Finally, we explore the use of positional encoding, which was not used originally during our main solution and the previous experiments, as initial testing indicated a slight decrease in performance. To confirm these initial findings we conduct a more detailed analysis. The results of this study are presented in Table~\ref{table:positional}, which confirm our initial findings that positional encoding does not improve performance and in fact decreases performance by a considerable margin. Positional encoding may introduce additional complexity when handling inherently noisy EEG signals. This can particularly be a problem if the noise pattern is not consistent across time, which is often the case with EEG data

  \begin{table}[t]
        \caption{Sensitivity analysis on the scheduler configurations.}
        \label{table:LRS} 
        \small
        \setlength
        \tabcolsep{2pt}
        \begin{center}{
                \begin{tabular}{lllll}
                 \hline
                 Learning Rate & Gamma & Step Size & Accuracy & F1 score \\
                 \hline\hline
                  0.0001 & 250 & 0.75 & \textbf{70.28 (0.027)} & \textbf{71.17 (0.031)}  \\
                  0.00001 & 250 & 0.75 & 68.25 (0.036) & 60.42 (0.044)  \\
                  0.0001 & 100 & 0.5 & 72.24 (0.026) & 66.23 (0.038)  \\
                  0.0001 & 150 & 0.5 & 74.43 (0.014) & 71.12 (0.026) \\
                 \hline
                \end{tabular} 
                }
        \end{center}
\end{table}

\subsection{Sensitivity Analysis}

We conduct sensitivity analyses on a variety of learning rate scheduler configurations to determine the most effective configuration. These configurations and the results are described in Table~\ref{table:LRS}. For each learning rate scheduler configuration, there is an initial learning rate value, a gamma value that represents the multiplicative factor by which the learning rate decreases for every step, and the step size which determines the frequency of epochs where the learning rate will decrease. The results indicate that a learning rate of 0.0001 and Gamma of 250 with a step size of 0.75 is the most effective configuration for the downstream EEG cognitive load classification task. With regard to architecture and pre-training datasets, we use the pipeline optimized through our ablation studies in this analysis.

\section{Conclusion and Future Work} \label{Conclusion&FutureWork}
We presented a new approach for cognitive load classification from EEG using transformer architectures. Our method used masked encoding of tokenized features for pre-training on emotion datasets. This was followed by transferring the model for downstream classification of cognitive load. Our method demonstrates the potential of self-supervised pre-training through masked autoencoding in combination with cross-domain transfer learning as a promising approach for classifying cognitive load from EEG data. This approach surpasses conventional single-stage fully supervised learning when classifying EEG cognitive load, by an average increase of 8.33\% in accuracy and 11.33\% in F1 score. Detailed ablation and sensitivity studies demonstrated the impact of different components and variants of our method. Our new approach makes a valuable contribution to the advancement in the field of cognitive load analysis, which currently remains relatively under-explored in the context of EEG signals. The insights gained from this study can guide future research aimed at improving the detection of high levels of cognitive load in high-risk situations, thereby enhancing human performance and safety.

For future work, to leverage unlabeled pre-training data (e.g., from emotion datasets) together with labeled downstream data (e.g., from cognitive load datasets), semi-supervised learning could be explored. Semi-supervised learning has shown promising results in the area of emotion recognition from EEG; however, its use in cross-domain transfer learning has remained unexplored in the context of cognitive load classification. Moreover, given the difficulty of generating accurate labels for cognitive load datasets, the notion of partial label learning could be studied. This notion has been recently explored for emotion recognition from EEG, but is yet to be applied for cognitive load classification.

\bibliographystyle{ACM-Reference-Format}
\bibliography{bib}


\end{document}